# Optimized Fuzzy Logic Approach with the IEEE Key Gas Method for Diagnosing Power Transformer Faults Using Dissolved Gas Analysis

Kim Anh Nguyen
Faculty of Electrical Engineering
The University of Danang - University of Science and Technology
Da Nang, Vietnam
https://orcid.org/0000-0003-3408-847X
Corresponding author: Kim Anh Nguyen (nkanh@dut.udn.vn)

Huy Hoang Le
Faculty of Electrical Engineering
The University of Danang - University of Science and Technology
Da Nang, Vietnam
105200406@sv1.dut.udn.vn

Ba Tu Phung
Faculty of Electrical Engineering
The University of Danang - University of Science and Technology
Da Nang, Vietnam
105200478@sv1.dut.udn.vn

***Abstract*—Reliable transformer fault diagnosis is essential for maintaining power system stability. The IEEE Key Gas Method (KGM), a widely utilized approach in Dissolved Gas Analysis (DGA), exhibits limitations in addressing ambiguous data and ensuring high diagnostic accuracy. This study presents An enhanced model combining Fuzzy Logic with the IEEE Key Gas Method (FL-KGM) that introduces refined membership functions, optimized fuzzy rule sets, and a novel separation of CO and $CO_2$ to eliminate diagnostic inconsistencies. By leveraging multidimensional gas ratio analysis and an adaptive classification framework, FL-KGM delivers superior fault identification and classification. Experimental validation utilizing real-world datasets demonstrates that FL-KGM achieves up to 98.6% accuracy, significantly outperforming KGM and other FL-based approaches. These findings elucidate the potential of FL-KGM in advancing transformer monitoring, enabling intelligent fault detection, and enhancing predictive maintenance strategies in modern power systems.**

## I. INTRODUCTION (*HEADING 1*)

This Transformers are critical components in electrical systems, playing a pivotal role in the transmission and distribution of energy. Failures in transformers not only result in significant economic losses but also severely impact the stability and reliability of power systems. Consequently, early detection and accurate diagnosis of potential faults have become essential requirements to ensure safe and efficient operation [1]. DGA is widely recognized as an effective tool for early transformer fault detection. The KGM, based on IEEE standards, has been extensively applied due to its ability to simplify the analysis of gases generated in insulating oil [2], [3]. However, traditional methods such as KGM rely on fixed thresholds, which pose limitations in handling complex, ambiguous, or outlier data [4]. Fuzzy Logic (FL) has demonstrated efficacy in addressing uncertainty, particularly in complex data analysis systems such as DGA. When integrated into the KGM, FL enhances analytical capabilities and improves diagnostic accuracy. Nevertheless, conventional FL systems still face challenges related to flexibility, scalability, and substantial reliance on expert experience [5]. This study introduces an improved FL-KGM, which incorporates significant modifications to enhance diagnostic precision. The primary contribution of this research lies in the independent evaluation of CO and $CO_2$, rather than considering them collectively. This adjustment prevents CO from disproportionately influencing the diagnostic process, leading to more accurate fault identification, especially in insulation degradation and thermal condition assessment [6]. Additionally, a completely redesigned FL system for key gas analysis has been developed using real-world datasets, ensuring a more data-driven and adaptive fault classification model. The proposed FL-KGM is validated through simulations in Matlab/Simulink, demonstrating superior accuracy and reliability compared to traditional and existing fuzzy-based methods, reinforcing its practical applicability in transformer fault diagnosis.

## II. PROPOSED FUZZY LOGIC-BASED IEEE KEY GAS MODEL

To develop an effective fault diagnosis model, this study focuses on two core issues: (i) examining traditional key gas fault diagnosis methods based on evaluating the percentage ratios of gases generated during transformer operation and (ii) exploring FL fundamentals while enhancing FL-KGM.

### *A. IEEE Key Gas Method and Limitations*

DGA is regarded as an effective technique for diagnosing potential faults in transformers by analyzing gases dissolved in insulating oil [2], [3], [4]. Gases such as Hydrogen ($H_2$), Methane ($CH_4$), Ethane ($C_2H_6$), Ethylene ($C_2H_4$), Acetylene ($C_2H_2$), Carbon Monoxide (CO), and Carbon Dioxide ($CO_2$) are produced during phenomena such as partial discharge (PD), overheating (T), or discharge (D) [6], [7]. Acceptable concentration limits for these fault-related gases can differentiate between normal and abnormal operating conditions and serve as warning signals that trigger in-depth analysis using DGA-based diagnostic methods. This enables a clear distinction between faulty and nonfaulty states. Fig. 1 illustrates the permissible limits proposed in literature [2], [7], [8], [9], [10], [11], [12].

Among the various DGA-based techniques, key gas analysis has emerged as a fundamental approach due to its capacity to correlate specific gas formations with distinct fault types. This method traces its origins to the mid-1950s when Howe [13] initially investigated the diagnostic significance of gases such as CO, $C_2H_2$, and $H_2$ in Buchholz relays. His findings established the foundation for the development of structured key gas-based diagnostic methods, which have subsequently evolved to enhance fault identification accuracy. Over time, several methodologies have been introduced, including the IEEE Key Gas method [2], the LCIE method [14], and the approach developed by California State University, Sacramento (CSUS) [12]. Additional models such as the Total Dissolved Combustible Gas (TDCG) method [7], and the key gas model proposed by Muller et al. [6] further exemplify the continued advancement of key gas analysis in transformer fault detection and condition monitoring.

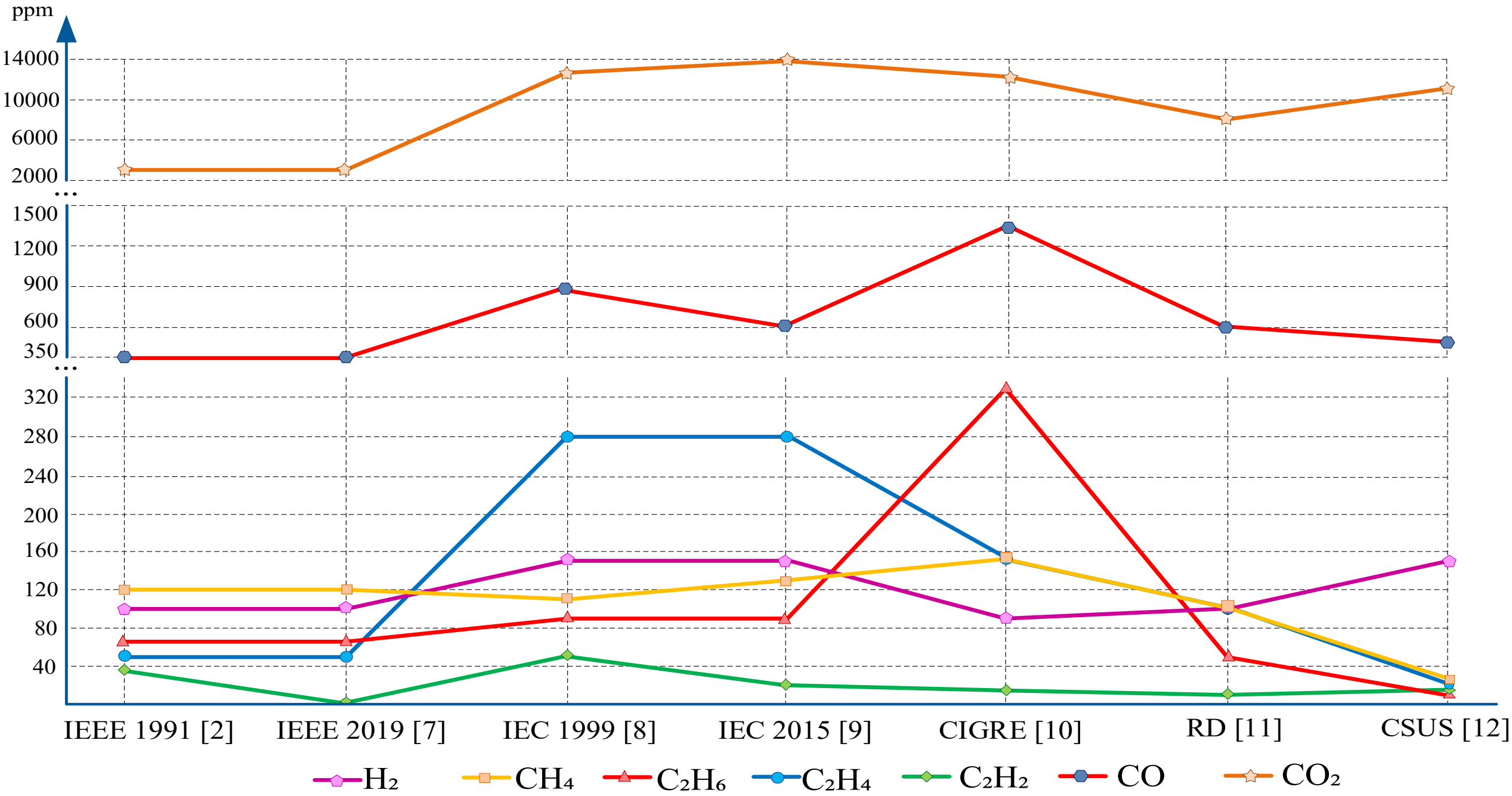

Fig. 1. Permissible fault gas levels according to various standards.

Building upon these advancements, the IEEE Key Gas method emerged as one of the most widely adopted DGA techniques. Originally developed by Doble Engineering Company in 1973 and subsequently refined by David Pugh in 1974 [15], this approach became an industry standard, officially recognized in IEEE Std C57.104. It operates by analyzing the ratios of gases generated under thermal and electrical stresses within transformers, with a primary focus on hydrogen, hydrocarbons, and carbon monoxide—key indicators of oil and insulation degradation. This methodology classifies fault conditions based on predefined gas concentration thresholds, with Table I summarizing the key diagnostic gases and their associated fault types, while Table II outlines the characteristic gas ratios used for fault identification, derived from extensive industry experience [7].

TABLE I. KG GENERATED BY FAULT TYPE [7]

| Key gas | Fault type | Typical component ratio |
|---|---|---|
| $C_2H_4$ | Overheating (oil) | Mainly $C_2H_4$; small amounts of $C_2H_6$, $CH_4$, and $H_2$; trace amounts of $C_2H_2$ occurring in high-temperature fault scenarios. |
| CO | Overheating (paper and oil) | Primarily CO; significantly smaller amounts of hydrocarbon gases (primarily $C_2H_4$) along with lesser amounts of $C_2H_6$, $CH_4$, and $H_2$. |
| $H_2$ | Partial discharge | Mainly $H_2$; small amounts of $CH_4$; trace amounts of $C_2H_4$ and $C_2H_6$. |
| $H_2$ & $C_2H_2$ | Discharge | Mainly $H_2$ and $C_2H_2$; small amounts of $CH_4$, $C_2H_4$, and $C_2H_6$; CO may also be present if cellulose is involved. |

TABLE II. KG GENERATED BY FAULT TYPE BASED ON PERCENTAGE RATIO [7]

| Fault type | $H_2$ | $CH_4$ | $C_2H_6$ | $C_2H_4$ | $C_2H_2$ | CO |
|---|---|---|---|---|---|---|
| Partial discharge | 85% | 13% | 1% | 1% | - | - |
| Discharge | 60% | 5% | 2% | 3% | 30% | - |
| Overheating (oil) | 2% | 16 | 19% | 63% | - | - |
| Overheating (paper and oil) | - | - | - | - | - | 92% |

Despite its widespread application, the IEEE Key Gas method has notable limitations. When integrated into software-based diagnostic systems, it exhibits a high rate of misclassification or inconclusive results, reaching up to 50% in some instances. Even with expert manual interpretation, the error rate remains at approximately 30% [7], highlighting the method's inherent challenges in handling complex or borderline cases. Moreover, in real-world transformer failures, gas profiles frequently deviate from the predefined Key Gas categories, leading to diagnostic inconsistencies. A significant issue is the use of CO as a direct indicator of insulation paper degradation, which can occasionally result in misdiagnoses [6]. These limitations underscore the necessity for more sophisticated diagnostic techniques, such as fuzzy logic-based enhancements, to improve reliability and decision-making in transformer fault detection.

### *B. Proposed FL-KGM Model*

Artificial Intelligence (AI) and soft computing effectively address uncertainties, particularly in transformer fault diagnosis [3], [4]. Fuzzy logic, proposed by Zadeh in 1965, is a crucial tool in DGA because of its ability to handle ambiguous data [16]. The operational framework comprises three steps: fuzzification, fuzzy computation, and defuzzification Fig. 2.

A fuzzy logic system infers fault conditions from gas data using fuzzy rules and defuzzification processes [17]. However, traditional methods face optimization limitations when dealing with large datasets and complex faults [18]. To address these challenges, recent studies have integrated fuzzy logic with machine learning techniques, such as fuzzy neural networks and adaptive fuzzy systems, enabling automated learning and the optimization of diagnostic rules [19].

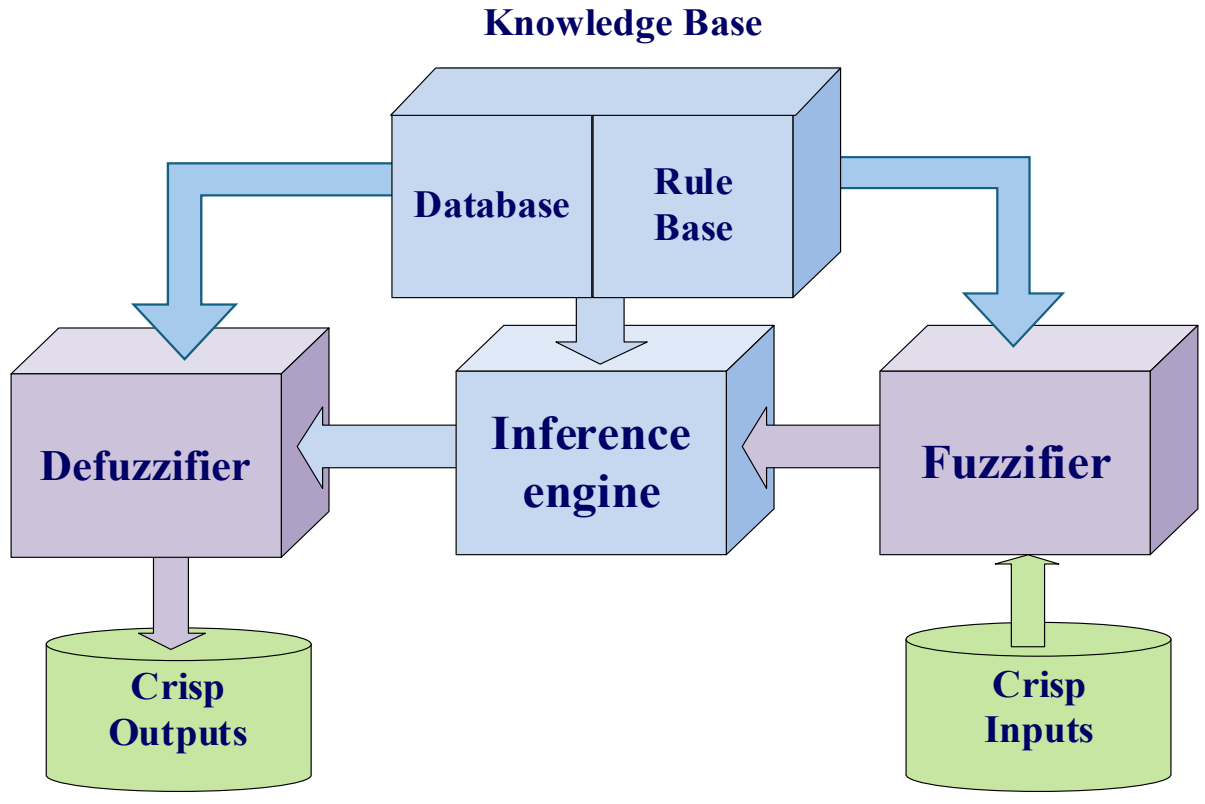

Fig. 2. Principle flowchart of fuzzy logic.

The KGM, as per IEEE standards, is simple and effective but has limitations in handling ambiguous data and exceptional cases, resulting in suboptimal diagnostic accuracy [2]. In the traditional KGM [7], the percentage ratios of six gases ($H_2$, $CH_4$, $C_2H_2$, $C_2H_4$, $C_2H_6$, and CO) were utilized. However, the presence of CO can skew diagnostic outcomes and reduce accuracy; detailed data and analysis are presented later. This paper proposes an improved FL system with key modifications. Instead of using all six gases, this study used five gases, excluding CO, and added a criterion for identifying thermal faults in insulating paper based on the $CO_2$/CO ratio [7]. In addition, fuzzy rules have been optimized to handle complex scenarios and outliers using real-world data. The membership functions of the FL system are refined to enhance accuracy and minimize ambiguity in fault classification. Mamdani method and centroid defuzzification were employed to ensure balanced and consistent diagnostic results.

The FL-KGM is developed based on diagnostic rules from studies [2], [7], with redesigned membership functions for input gases ($H_2$, $CH_4$, $C_2H_2$, $C_2H_4$, $C_2H_6$). These functions are constructed according to the value ranges in Table II and are enhanced to address the limitations of previous research [20]. To optimize the performance, membership functions were adjusted in shape and slope, with value ranges redivided more rationally, as shown in Table III and Fig. 3. Table III depicts the output membership function set comprising five functions: F0-F4, representing no fault (N), thermal fault (T), partial discharge (PD), discharge (D), and undefined (UD), with values from 0 to 10. Fig. 3 presents also the FL model integrated with the KGM, featuring 81 decision rules based on five input gas percentage ratios, expressed as FL membership sets.

By refining membership functions, the FL-KGM model achieves a significant improvement in diagnostic accuracy compared to the traditional key gas approach. To further enhance its performance and adaptability, this study introduces an optimized set of fuzzy rules, as presented in Table IV. These rules effectively address the issue of overlapping thresholds, wherein gas ratios fall into transitional zones between fault categories. In contrast to conventional methods that rely on rigid classification boundaries, the improved fuzzy system dynamically interprets these values, ensuring more accurate and consistent fault identification. A key enhancement of the new rule set is its capacity to account for subtle variations in gas composition that might otherwise result in misdiagnoses. Traditional key gas-based methods often encounter difficulties with cases where gas ratios fall marginally outside predefined limits, leading to inconclusive outcomes. By integrating multiple gas ratios into a comprehensive decision-making framework, the FL-KGM model minimizes such uncertainties, thereby improving reliability across diverse operating conditions. This multidimensional analysis facilitates a more thorough assessment of transformer health, ensuring the system remains effective even when confronted with atypical gas patterns or borderline cases.

TABLE III. OUTPUT MEMBERSHIP FUNCTIONS OF FL-KGM

| Fault | Result | Auxiliary criterion | Diagnosed fault |
|---|---|---|---|
| F0 | $D \leq 2$ | | Normal aging (N) |
| F1 | $2 < D \leq 4$ | $CO_2/CO \geq 3.0$ | Local overheating not involving cellulose (T) |
| | | $CO_2/CO < 3.0$ | Local overheating involving cellulose (T + C) |
| F2 | $4 < D \leq 6$ | | Partial Discharge (PD) |
| F3 | $6 < D \leq 8$ | | Discharge (D) |
| F4 | $8 < D \leq 10$ | | Undeterminate (UD) |

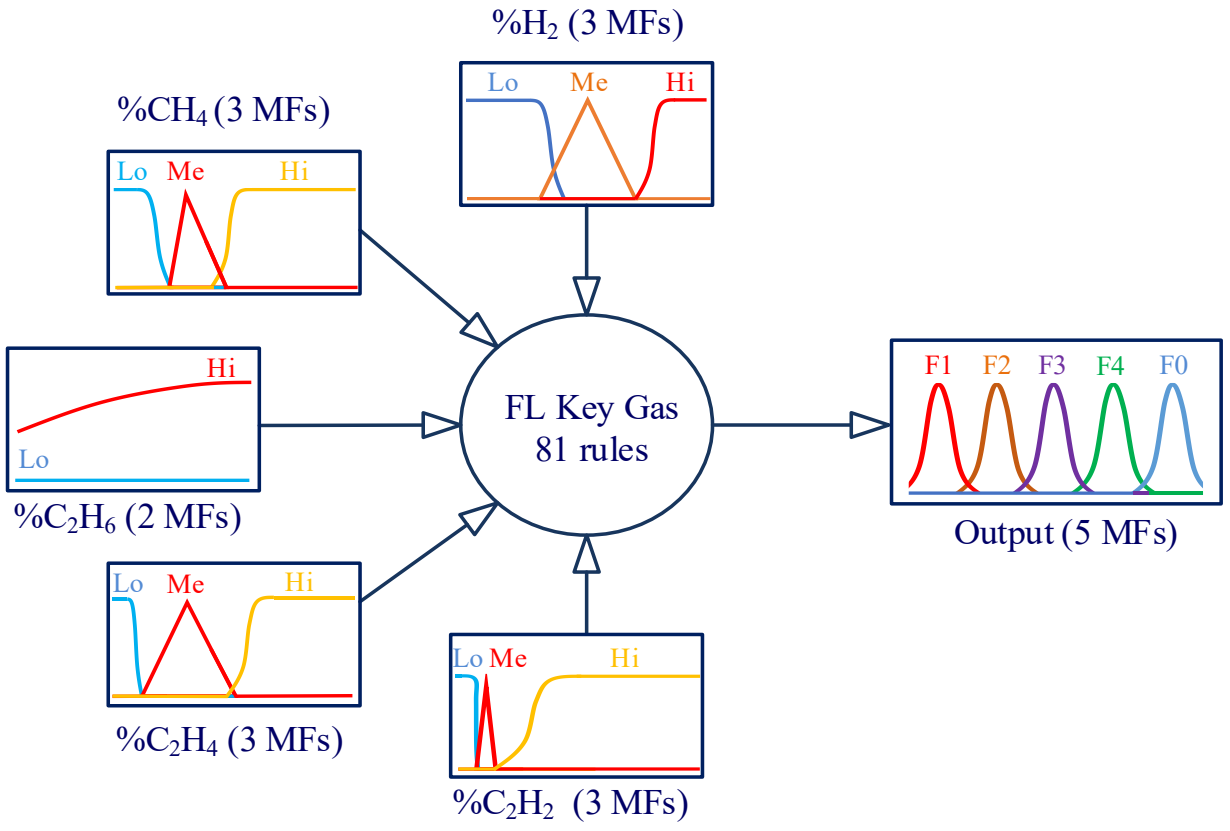

Fig. 3. Optimized membership functions of the propoed FL-KGM model.

TABLE IV. 81 RULES OF THE PROPOSED FL-KGM MODEL

| Rule | $\%H_2$ | $\%CH_4$ | $\%C_2H_6$ | $\%C_2H_4$ | $\%C_2H_2$ | Output |
|---|---|---|---|---|---|---|
| 1 | Lo | Lo | Hi | Lo | Me | F3 |
| 2 | Hi | Lo | Hi | Lo | Me | F3 |
| 3 | Me | Lo | Hi | Lo | Me | F3 |
| 4 | Lo | Me | Hi | Lo | Me | F3 |
| 5 | Hi | Me | Hi | Lo | Me | F3 |
| 6 | Me | Me | Hi | Lo | Me | F3 |
| 7 | Lo | Hi | Hi | Lo | Me | F3 |
| 8 | Hi | Hi | Hi | Lo | Me | F3 |
| 9 | Me | Hi | Hi | Lo | Me | F3 |
| 10 | Lo | Lo | Hi | Me | Me | F3 |
| 11 | Hi | Lo | Hi | Me | Me | F3 |
| 12 | Me | Lo | Hi | Me | Me | F1 |
| 13 | Lo | Me | Hi | Me | Me | F1 |
| 14 | Hi | Me | Hi | Me | Me | F3 |
| 15 | Me | Me | Hi | Me | Me | F3 |
| 16 | Lo | Hi | Hi | Me | Me | F3 |
| 17 | Hi | Hi | Hi | Me | Me | F3 |
| 18 | Me | Hi | Hi | Me | Me | F3 |
| 19 | Lo | Lo | Hi | Hi | Me | F1 |
| 20 | Hi | Lo | Hi | Hi | Me | F3 |
| 21 | Me | Lo | Hi | Hi | Me | F3 |
| 22 | Lo | Me | Hi | Hi | Me | F1 |
| 23 | Hi | Me | Hi | Hi | Me | F3 |
| 24 | Me | Me | Hi | Hi | Me | F3 |

| Rule | %$H_2$ | %$CH_4$ | %$C_2H_6$ | %$C_2H_4$ | %$C_2H_2$ | Output |
|---|---|---|---|---|---|---|
| 25 | Lo | Hi | Hi | Hi | Me | F1 |
| 26 | Hi | Hi | Hi | Hi | Me | F3 |
| 27 | Me | Hi | Hi | Hi | Me | F3 |
| 28 | Lo | Lo | Hi | Lo | Lo | F1 |
| 29 | Hi | Lo | Hi | Lo | Lo | F2 |
| 30 | Me | Lo | Hi | Lo | Lo | F1 |
| 31 | Lo | Me | Hi | Lo | Lo | F1 |
| 32 | Hi | Me | Hi | Lo | Lo | F2 |
| 33 | Me | Me | Hi | Lo | Lo | F1 |
| 34 | Lo | Hi | Hi | Lo | Lo | F1 |
| 35 | Hi | Hi | Hi | Lo | Lo | F4 |
| 36 | Me | Hi | Hi | Lo | Lo | F1 |
| 37 | Lo | Lo | Hi | Me | Lo | F1 |
| 38 | Hi | Lo | Hi | Me | Lo | F1 |
| 39 | Me | Lo | Hi | Me | Lo | F1 |
| 40 | Lo | Me | Hi | Me | Lo | F1 |
| 41 | Hi | Me | Hi | Me | Lo | F1 |
| 42 | Me | Me | Hi | Me | Lo | F1 |
| 43 | Lo | Hi | Hi | Me | Lo | F1 |
| 44 | Hi | Hi | Hi | Me | Lo | F4 |
| 45 | Me | Hi | Hi | Me | Lo | F1 |
| 46 | Lo | Lo | Hi | Hi | Lo | F1 |
| 47 | Hi | Lo | Hi | Hi | Lo | F4 |
| 48 | Me | Lo | Hi | Hi | Lo | F4 |
| 49 | Lo | Me | Hi | Hi | Lo | F1 |
| 50 | Hi | Me | Hi | Hi | Lo | F4 |
| 51 | Me | Me | Hi | Hi | Lo | F4 |
| 52 | Lo | Hi | Hi | Hi | Lo | F1 |
| 53 | Hi | Hi | Hi | Hi | Lo | F4 |
| 54 | Me | Hi | Hi | Hi | Lo | F4 |
| 55 | Lo | Lo | Hi | Lo | Hi | F3 |
| 56 | Hi | Lo | Hi | Lo | Hi | F3 |
| 57 | Me | Lo | Hi | Lo | Hi | F3 |
| 58 | Lo | Me | Hi | Lo | Hi | F3 |
| 59 | Hi | Me | Hi | Lo | Hi | F3 |
| 60 | Me | Me | Hi | Lo | Hi | F3 |
| 61 | Lo | Hi | Hi | Lo | Hi | F3 |
| 62 | Hi | Hi | Hi | Lo | Hi | F3 |
| 63 | Me | Hi | Hi | Lo | Hi | F3 |
| 64 | Lo | Lo | Hi | Me | Hi | F3 |
| 65 | Hi | Lo | Hi | Me | Hi | F3 |
| 66 | Me | Lo | Hi | Me | Hi | F3 |
| 67 | Lo | Me | Hi | Me | Hi | F3 |
| 68 | Hi | Me | Hi | Me | Hi | F3 |
| 69 | Me | Me | Hi | Me | Hi | F3 |
| 70 | Lo | Hi | Hi | Me | Hi | F3 |
| 71 | Hi | Hi | Hi | Me | Hi | F3 |
| 72 | Me | Hi | Hi | Me | Hi | F3 |
| 73 | Lo | Lo | Hi | Hi | Hi | F3 |
| 74 | Hi | Lo | Hi | Hi | Hi | F3 |
| 75 | Me | Lo | Hi | Hi | Hi | F3 |
| 76 | Lo | Me | Hi | Hi | Hi | F3 |
| 77 | Hi | Me | Hi | Hi | Hi | F3 |
| 78 | Me | Me | Hi | Hi | Hi | F3 |
| 79 | Lo | Hi | Hi | Hi | Hi | F3 |
| 80 | Hi | Hi | Hi | Hi | Hi | F3 |
| 81 | Me | Hi | Hi | Hi | Hi | F3 |

Fig. 4 presents the diagnostic framework for evaluating transformer conditions utilizing the FL-KGM model, integrating key gas analysis with $CO_2$/CO ratio assessment to enhance accuracy and reliability in fault detection. The process commences with an initial evaluation of the transformer's operational status. If the parameters indicate normal operation, no further action is necessitated. However, if abnormalities are detected, the FL-KGM analyzes dissolved gas ratios to determine fault type and severity.

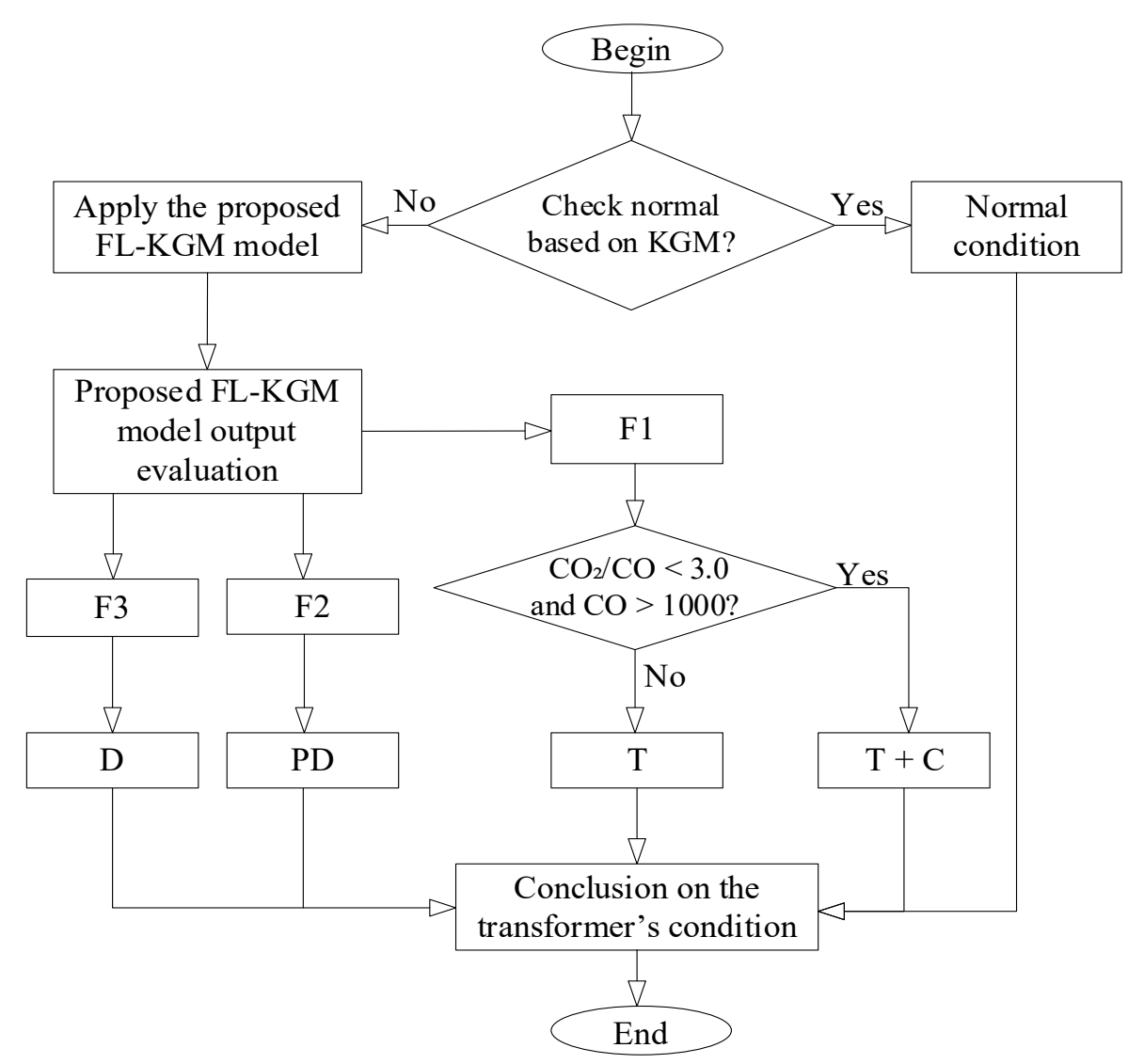

Fig. 4. Diagnostic process flowchart of the proposed FL-KGM model.

For cases classified as F1 (thermal faults), the system subsequently examines the $CO_2$/CO ratio—a critical indicator of insulation paper integrity [7]. If the ratio is below three, the insulation remains intact; otherwise, the system indicates potential risks associated with insulation degradation or abnormal temperature conditions. The process concludes with a comprehensive diagnosis, facilitating timely maintenance and corrective actions. By integrating multiple diagnostic criteria, this approach not only improves fault classification precision but also enhances adaptability for practical transformer monitoring and management.

## III. Simulation and Evaluation of the FL-KGM Model

To evaluate the effectiveness of the improved FL system, the KGM was implemented and tested through simulations using the MATLAB/Simulink platform. The simulation model utilizes a real-world DGA dataset to assess the fault identification capabilities and compare the performance of the traditional method and enhanced fuzzy logic system. The simulation results highlight the extent to which the proposed FL system improves diagnostic accuracy.

### *A. MATLAB/Simulink-Based Simulation Model*

The model comprises key components, including a normal-state verification block, data preprocessing block, FL block, $CO_2$/CO ratio analysis block, and result display block. The data preprocessing block extracts gas concentration values ($H_2$, $CH_4$, $C_2H_6$, $C_2H_4$, $C_2H_2$, CO, $CO_2$) from the DGA sample and calculates percentage ratios utilizing the KGM. Upon detection of an abnormality, the processed data is transmitted to the FL block, where fuzzy sets and inference rules (see Table IV) are applied using MATLAB's Fuzzy Logic Toolbox. The outputs are subsequently defuzzified and analyzed through the $CO_2$/CO ratio to assess insulation and thermal conditions. Finally, the results are displayed, providing a comprehensive diagnosis of the transformer's condition. A detailed illustration of the complete simulation model is presented in Fig. 5.

### *B. Simulation outcomes and assessment*

To visually assess the efficacy of the diagnostic methods, 15 DGA samples were randomly selected from the dataset,

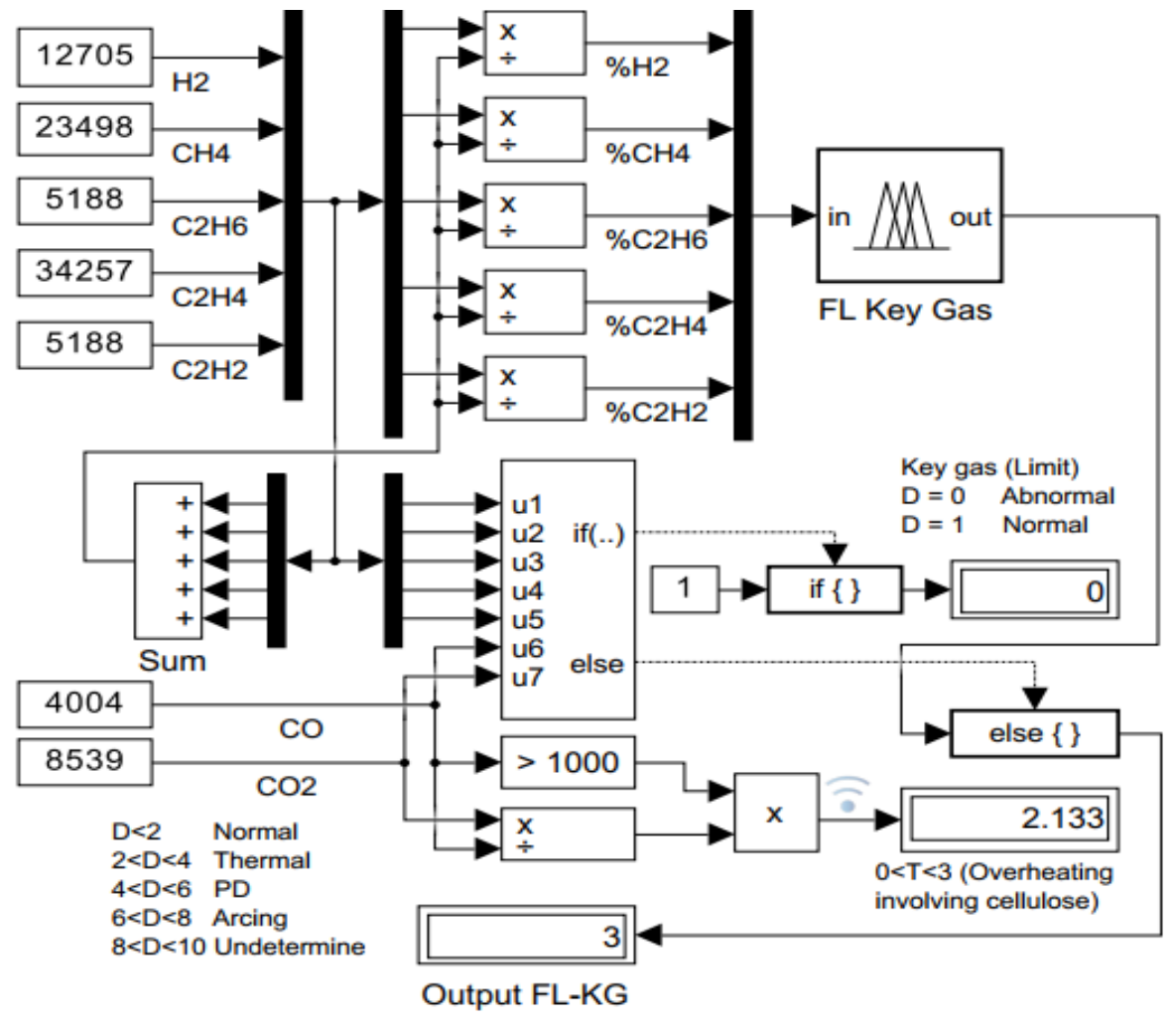

Fig. 5. Simulation of the proposed FL-KGM model using MATLAB/Simulink.

encompassing various fault conditions, including normal operation, PD, T, and D. The findings are detailed in Tables V and VI. Among the evaluated samples, the traditional KGM, when incorporating CO gas percentages, accurately identified faults in only 7 out of 15 instances, highlighting potential limitations in its diagnostic capability when CO concentration is considered. Notably, excluding CO gas percentages markedly enhanced accuracy to 12 out of 15, suggesting that certain gas ratios may introduce ambiguity in fault classification when used in isolation. Conversely, the proposed FL-KGM accurately diagnosed all 15 samples, even in complex scenarios such as insulating paper overheating, where traditional DGA-based methods often encounter difficulties due to overlapping fault signatures.

To thoroughly evaluate diagnostic reliability and generalizability, the we further assessed the three methods using 150 DGA samples for testing [17], [18], [20], [21], thereby ensuring robust statistical validation. Accuracy was measured to ascertain the consistency of each method across various fault categories, with the results compared to the conventional KGM as well as FL-enhanced versions of the IEC, RRM, and DRM methodologies, as detailed in [5]. The performance metrics are summarized in Table VI and visually depicted in Fig. 6.

TABLE V. ACCURACY COMPARISION OF THE KGM AND THE PROPOSED FL-KGM MODEL

| Method/ Model | Error Code | Number of samples | Correctly diagnosed samples | Correct diagnosis percentage | Accuracy |
|---|---|---|---|---|---|
| Traditional KGM | N | 34 | 33 | 97 | 62.6% |
| | T | 27 | 18 | 66.7 | |
| | T + C | 7 | 1 | 14.3 | |
| | PD | 8 | 7 | 87.5 | |
| | D | 74 | 35 | 47.3 | |
| Proposed FL-KGM | F0 | 34 | 34 | 100 | 98.6% |
| | F1 | 27 | 26 | 96.3 | |
| | F1 + C | 7 | 7 | 100 | |
| | F2 | 8 | 8 | 100 | |
| | F3 | 74 | 73 | 98.6 | |

An important observation derived from the results is that, when applied to a comprehensive dataset, the enhancements in accuracy become more evident. While traditional KGM demonstrates variable performance across different fault types, its accuracy remains below 85% in several scenarios, underscoring its vulnerability to misclassification, particularly in instances involving multiple overlapping fault signatures. In contrast, the FL-enhanced methods exhibit a distinct advantage, with all four FL models (FL-IEC, FL-RRM, FL-DTM, and FL-KGM) consistently surpassing their conventional counterparts. Notably, the proposed FL-KGM model distinguishes itself by achieving accuracy rates as shown in Table V for various fault categories, significantly exceeding not only the standard KGM but also other hybrid FL models, which generally do not surpass 90% accuracy. This superior performance can be attributed to the optimized fuzzy membership functions, expanded rule sets, and the integration of auxiliary criteria such as the $CO_2/CO$ ratio, which collectively enhance the model's capability to differentiate between similar fault conditions with high precision.

Furthermore, the proposed FL-KGM method demonstrates exceptional stability across a range of fault scenarios, highlighting its robustness against variations in gas concentrations and complex failure patterns. In contrast to conventional techniques, which frequently necessitate manual threshold adjustments for different transformer operating conditions, the FL-KGM model dynamically adapts to diverse fault profiles, thereby ensuring consistent performance in real-world applications. This adaptability is essential for practical deployment in transformer condition monitoring, where variations in oil aging, environmental factors, and load conditions can significantly influence diagnostic outcomes.

TABLE VI. ENHANCIED DIAGNOSTIC ACCURACY USING THE PROPOSED FL-KGM MODEL

| Sample | Input Data | | | | | | | Traditional KGM | | FL-KGM | Actual fault diagnosis |
|---|---|---|---|---|---|---|---|---|---|---|---|
| | $H_2$ | $CH_4$ | $C_2H_6$ | $C_2H_4$ | $C_2H_2$ | CO | $CO_2$ | Presence of %CO | Absence of %CO | | |
| 1 | 7.4 | 8.4 | 6.3 | 3.8 | 0 | 0.1 | 0.05 | N | N | F0 | Normal |
| 2 | 12 | 10.7 | 6.9 | 3.7 | 0 | 0.17 | 0.03 | N | N | F0 | Normal |
| 3 | 360 | 610 | 9 | 260 | 259 | 12000 | 74200 | UD | UD | F1 | Thermal |
| 4 | 1 | 8 | 6 | 100 | 6 | 300 | 5130 | UD | T | F1 | Thermal |
| 5 | 960 | 4000 | 6 | 1560 | 1290 | 15800 | 50300 | UD | T | F1 | Thermal |
| 6 | 480 | 1075 | 0 | 1132 | 298 | 464 | 1000 | T | T | F1+C | Thermal (T +C) |
| 7 | 12705 | 23498 | 5188 | 34257 | 5188 | 4004 | 8539 | T | T | F1+C | Thermal (T +C) |
| 8 | 48 | 610 | 0 | 10 | 29 | 1900 | 970 | UD | T | F1+C | Thermal (T +C) |
| 9 | 92600 | 10200 | 0.01 | 0.01 | 0.01 | 6400 | 103151 | UD | PD | F2 | PD |
| 10 | 32930 | 2397 | 157 | 0.01 | 0.01 | 313 | 560 | PD | PD | F2 | PD |
| 11 | 37800 | 1740 | 249 | 8 | 8 | 56 | 197 | PD | PD | F2 | PD |
| 12 | 120 | 25 | 1 | 8 | 40 | 500 | 1600 | UD | D | F3 | Discharge |
| 13 | 20000 | 13000 | 1850 | 29000 | 57000 | 2600 | 2430 | T | D | F3 | Discharge |
| 14 | 4419 | 3564 | 668 | 2861 | 2025 | 909 | 9082 | UD | UD | F3 | Discharge |
| 15 | 810 | 580 | 111 | 570 | 490 | 1100 | 6800 | UD | UD | F3 | Discharge |

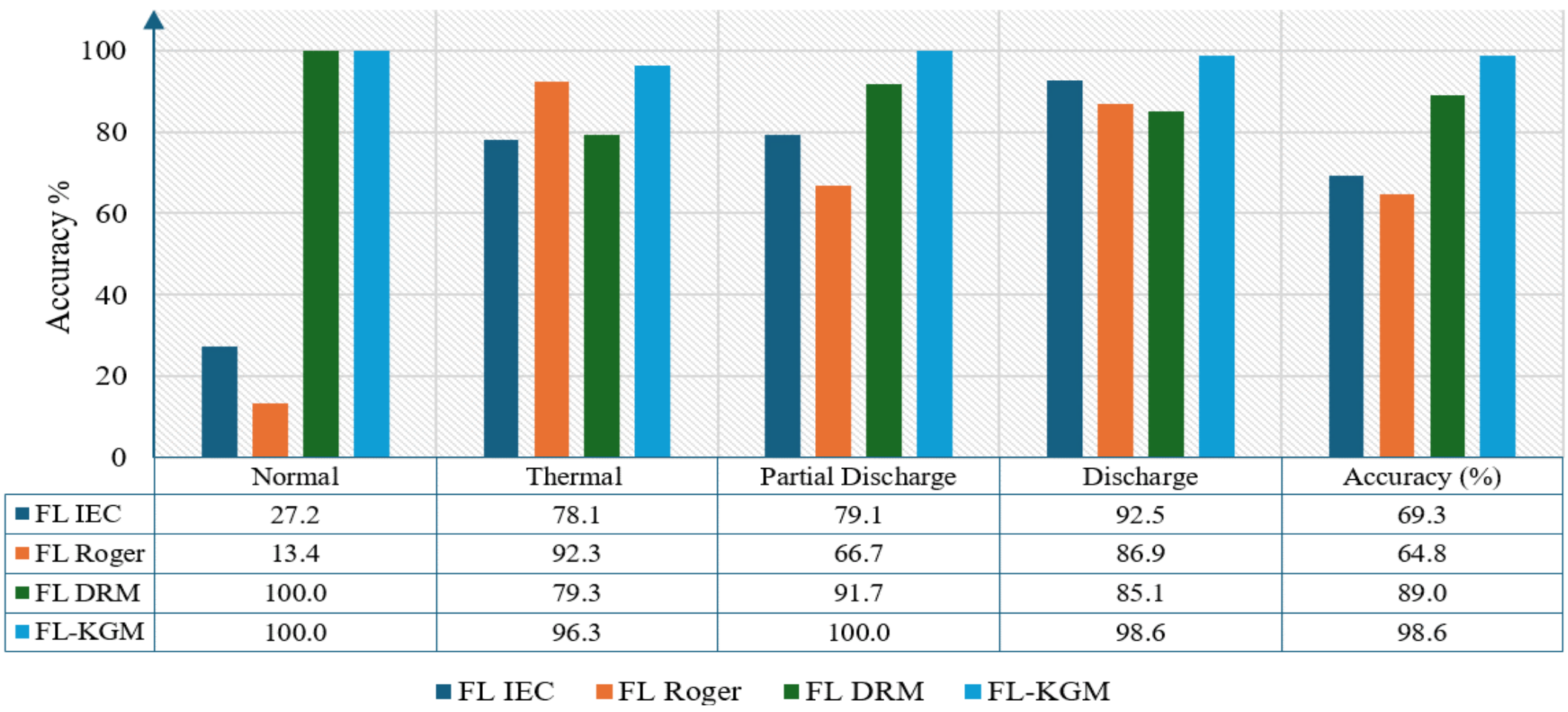

| | Normal | Thermal | Partial Discharge | Discharge | Accuracy (%) |
|---|---|---|---|---|---|
| FL IEC | 27.2 | 78.1 | 79.1 | 92.5 | 69.3 |
| FL Roger | 13.4 | 92.3 | 66.7 | 86.9 | 64.8 |
| FL DRM | 100.0 | 79.3 | 91.7 | 85.1 | 89.0 |
| FL-KGM | 100.0 | 96.3 | 100.0 | 98.6 | 98.6 |

Fig. 6. Accuracy and consistency of FL-KGM versus FL models with traditional methods in Gouda (2021) [5].

## IV. Conclusion

This study presents an enhanced FL model integrated with the traditional KGM, marking a significant advancement in transformer fault diagnosis. Key improvements, including optimized membership functions, expanded fuzzy rule sets, and the incorporation of auxiliary criteria such as the $CO_2/CO$ ratio, have augmented the model's capacity to process complex data while substantially improving fault classification accuracy. Experimental results indicate that the improved FL model achieves up to 98.6% accuracy, surpassing not only the KGM but also several state-of-the-art approaches. Notably, the model excels in identifying complex faults, such as partial discharge, thermal faults, and electrical arcing, demonstrating superior efficiency and stability. Building upon these findings, future research will focus on integrating diagnostic information with a Health Index based on DGA to develop predictive maintenance strategies for large power transformers. By leveraging AI-driven predictive analytics and real-time monitoring, the proposed system aims to enhance reliability, optimize maintenance schedules, and minimize unexpected failures, thereby setting the stage for broader automation and smart grid applications.

## References

[1] V. A. Thiviyanathan, P. J. Ker, Y. S. Leong, F. Abdullah, A. Ismail, and M. Zaini Jamaludin, "Power transformer insulation system: A review on the reactions, fault detection, challenges and future prospects," Alexandria Eng. J., vol. 61, no. 10, pp. 7697–7713, 2022, doi: 10.1016/j.aej.2022.01.026.

[2] Y. Siregar and T. J. Hartanto Lumbanraja, "Analysis of interference methods on transformers based on the results of dissolved gas analysis tests," Int. J. Electr. Comput. Eng., vol. 13, no. 4, pp. 3672–3685, 2023, doi: 10.11591/ijece.v13i4.pp3672-3685.

[3] S. A. Wani, A. S. Rana, S. Sohail, O. Rahman, S. Parveen, and S. A. Khan, “Advances in DGA based condition monitoring of transformers: A review,” Renewable Sustainable Energy Rev., vol. 149, pp. 111347, 2021.

[4] A. Wajid, A. U. Rehman, S. Iqbal, M. Pushkarna, S. M. Hussain, H. Kotb, and I. Zaitsev, “Comparative performance study of dissolved gas analysis (DGA) methods for identification of faults in power transformer,” Int. J. Energy Res., vol. 2023.1, pp. 9960743, 2023.

[5] O. E. Gouda, S. S. M. Ghoneim, and S. H. El-Hoshy, “Enhancing the Diagnostic Accuracy of DGA Techniques Based on IEC-TC10 and Related Databases,” IEEE Access, vol. 9, pp. 118031–118041, 2021.

[6] A. Nanfak, C. Hubert Kom, M. G. Ngaleu, F. Meghnefi, E. Samuel, and I. Fofana, "Traditional fault diagnosis methods for mineral oil-immersed power transformer based on dissolved gas analysis: Past, present and future," IET Nanodielectrics, vol. 7, no. 3, pp. 97–130, 2024, doi: 10.1049/nde2.12082.

[7] PES Transformers Committee, “IEEE guide for the interpretation of gases generated in mineral oil-immersed transformers,” IEEE Std C57.104, pp. 40–42.

[8] IEC Standard 60599,” Mineral Oil-Impregnated Electrical Equipment in Service—Guide to the Interpretation of Dissolved and Free Gases Analysis,” Int. Electrotech. Commission, Geneva, Switzerland, 1999.

[9] IEC Standard 60599, “Mineral Oil-Impregnated Electrical Equipment in Service—Guide to the Interpretation of Dissolved and Free Gases Analysis,” Int. Electrotech. Commission, Geneva, Switzerland, 2015.

[10] M. K. Ngwenyama and M. N. Gitau, "Discernment of transformer oil stray gassing anomalies using machine learning classification techniques," Sci. Rep., vol. 14, no. 1, 2024, doi: 10.1038/s41598-023-50833-7.

[11] V. D. Bitney, M. A. Moiseev, and V. Y. Ulyanov, “The need for using various interpretation methods for power transformer DGA results,” Mar. 2023, pp. 1–6. doi: 10.1109/reepe57272.2023.10086721.

[12] G. K. Irungu, A. O. Akumu, and J. L. Munda, “Transformer condition assessment for maintenance ranking: A comparison of three standards and different weighting techniques,” Sep. 2017, vol. c57, pp. 1043–1048. doi: 10.1109/afrcon.2017.8095626.

[13] V. H. Howe, L. Massey, and A. C. M. Wilson, “The identity and significance of gases collected in buchholz protectors,” Metropolitan-Vickers Electr. Co., 1956.

[14] Working Group 1E5, “Gas in oil analysis for fault detection,” Electr. Res. Assoc., 1976.

[15] D. David, “Advances in fault diagnosis by combustible gas analysis,” In Proceedings of the Minutes of 41st International Conference of Doble Clients, pp. 101201–101208, 1974.

[16] L.A. Zadeh, “Fuzzy sets,”, Information and Control, vol. 8, no. 3, pp. 338–353, 1965.

[17] K. A. Nguyen, H. H. Le, B. T. Phung, h. V. Tran, “A fuzzy logic approach combined with IEC and Roger methods for power transformer fault diagnosis based on DGA,” Univ. Danang, J. Sci. Technol., 2025, in press.

[18] K. A. Nguyen, H. H. Le, B. T. Phung, h. V. Tran, “Enhancing data preprocessing layer for power transformer fault diagnosis based on dissolved gas analysis combining fuzzy logic and Duval triangle 1,” The 8th International Conference on Circuits, Systems and Simulation, Industrial University of Ho Chi Minh City, Vietnam, 2025, in press.

[19] S. A. Khan, M. D. Equbal, and T. Islam, “A comprehensive comparative study of DGA based transformer fault diagnosis using fuzzy logic and ANFIS models,” IEEE Trans. Dielectr. Electr. Insul., vol. 22, no. 1, pp. 590–596, 2015.

[20] A. Abu-Siada and S. Hmood, “A new fuzzy logic approach to identify power transformer criticality using dissolved gas-in-oil analysis,” Int. J. Electr. Power Energy Syst., vol. 67, pp. 401–408, 2014.

[21] M. Duval and A. Depabla, “Interpretation of gas-in-oil analysis using new IEC publication 60599 and IEC TC 10 databases,” IEEE Electr. Insul. Mag., vol. 17, no. 2, pp. 31–41, 2001.